\documentclass{article} 
\usepackage[preprint]{colm2026_conference}

\usepackage{microtype}
\usepackage{hyperref}
\usepackage{url}
\usepackage{booktabs}

\usepackage{lineno}

\definecolor{darkblue}{rgb}{0, 0, 0.5}
\hypersetup{colorlinks=true, citecolor=darkblue, linkcolor=darkblue, urlcolor=darkblue}

\usepackage{inconsolata}

\usepackage{enumerate}
\usepackage{amsthm}
\usepackage{graphicx}
\usepackage{epstopdf}
\usepackage{caption}
\usepackage{color}
\usepackage{enumitem}
\setlist[enumerate]{nosep} 
\usepackage{array}
\usepackage{tabularx}
\usepackage{multirow}
\usepackage{bm}
\usepackage{booktabs}
\usepackage{amssymb}
\usepackage{tabularx}
\newcolumntype{Y}{>{\centering\arraybackslash}X}

\usepackage{tcolorbox}
\tcbuselibrary{breakable}
\tcbuselibrary{skins}
\usepackage{caption}
\usepackage{subcaption}
\usepackage{listings}
\usepackage{wrapfig}
\usepackage{bbding}
\tcbuselibrary{skins}
\usepackage{colortbl}
\usepackage{fontawesome5}
\usepackage{authblk}
\usepackage{amsmath}
\usepackage{hyperref}
\usepackage{etoolbox}
\preto{\abstractkeywords}{\nolinenumbers} 

\title{Density-aware Soft Context Compression with Semi-Dynamic Compression Ratio}

\author[a]{Yijiong Yu}
\affil[a]{Oregon State University\\
    \texttt{\{yuyiji, huazheng.wang\}@oregonstate.edu}
}

\author[b]{Shuai Yuan}

\author[b]{Jie Zheng}

\author[a]{Huazheng Wang}
\author[b]{Ji Pei}

\affil[b]{DeepSolution\\
\texttt{research@deepsolution.chat}}

\begin{document}
\maketitle
\begin{abstract}
Soft context compression reduces the computational workload of processing long contexts in LLMs by encoding long context into a smaller number of latent tokens. However, existing frameworks apply uniform compression ratios, failing to account for the extreme variance in natural language information density. While adopting a density-aware dynamic compression ratio seems intuitive, empirical investigations reveal that models struggle intrinsically with operations parameterized by input dependent, continuous structural hyperparameters. To resolve this pitfall, we introduce Semi-Dynamic Context Compression framework. Our approach features a Discrete Ratio Selector, which predicts a compression target based on intrinsic information density and quantizes it to a predefined set of discrete compression ratios. It is efficiently jointly trained with the compressor on synthetic data, with the summary lengths as a proxy to create labels for compression ratio prediction. Extensive evaluations confirm that our density-aware framework, utilizing mean pooling as the backbone, consistently outperforms static baselines, establishing a robust Pareto frontier for context compression techniques. Our code, data and model weights are available at \href{https://github.com/yuyijiong/semi-dynamic-context-compress}{github}
\end{abstract}
\section{Introduction}

The computational bottleneck of processing long contexts in Large Language Models (LLMs) has driven significant interest in soft context compression~\citep{dai_pretraining_2025,cheng_xrag_2024,feldman_simple_2025,ge_-context_2024,li_500xcompressor_2024,liu_context_2025}. By transforming discrete token sequences into shorter, continuous latent representations, soft context compression drastically reduces both the time complexity and memory overhead associated with Key-Value (KV) caching. A classic soft compression pipeline typically consists of three components: an \textit{encoder} (often initialized from an LLM) that computes the compressed features from the original text, a \textit{converter} (usually a two-layer MLP) that aligns the dimensionality of the encoder's hidden states, and a \textit{decoder} (typically initialized from the same LLM) that accepts these compressed features as token embeddings in place of the original context to generate responses. 

However, a critical limitation persists across current soft compression frameworks: they apply fixed compression ratios uniformly, ignoring the extreme variance in natural language information density. Intuitively, a dense technical report requires a vastly different compression budget than a highly redundant conversational transcript. Existing methods typically offer a static set of alternative compression ratios (usually independently trained), forcing users to manually balance compression ratio and quality based on heuristics, which inevitably leads to either suboptimal efficiency or suboptimal quality for diverse contexts.

An intuitive solution is a fully dynamic compression mechanism, where the model automatically predicts and applies the optimal, continuous compression ratio or length based on the input text. However, our empirical investigations reveal a severe failure mode in this approach. We discover that LLMs struggle intrinsically with operations parameterized by input-dependent, continuous structural hyperparameters, such as allocating a highly variable, input-dependent number of compression tokens, leads to profound performance collapse. This collapse likely occurs because LLMs with finite parameters cannot adapt to an infinite spectrum of dynamically shifting sequence reductions, nor can limited training data sufficiently cover them.

To resolve this pitfall, we introduce the \textbf{Semi-Dynamic Context Compression} framework. This approach adapts to varying text densities while completely circumventing the unlearnable continuous hyperparameter problem. The core of "semi-dynamic" is a \textit{Discrete Ratio Selector (DRS)}: during inference, the model predicts a compression target based on the intrinsic information density of the context, but this continuous prediction is strictly quantized to a predefined set of fixed, discrete compression ratios (e.g., $2\times, 4\times, 8\times$). Furthermore, our method introduces a powerful user-facing advantage: by manipulating a simple $scale$ parameter at inference time, users can smoothly and continuously control the global compression aggressiveness across a corpus, which is more flexible than relying on rigidly fixed compression ratios.

To maximize computational efficiency, we design a single-stage joint training architecture that achieves both of the 2 tasks, ratio prediction and context encoding, within a single encoding pass. Also, we employ a single-stage training pipeline to eschew computationally expensive text-reconstruction pre-training in favor of a pure Supervised Fine-Tuning (SFT) paradigm driven by high-quality synthetic data. By utilizing the summary lengths generated by a teacher LLM as a proxy for information density, we create regression labels that effectively supervise the model's ratio prediction. While for supervising the context compression and decoding functions, we follow established practices by utilizing synthetic summarization, single and multi document QA tasks.

Crucially, while our semi-dynamic framework can technically enhance various feature extraction (from the output hidden states of the encoder) methods, our rigorous benchmarking finds that without heavy pre-training, appending trainable compression tokens, the prevailing method, is significantly outperformed even by simple \textit{mean-pooling}. So we choose mean-pooling as the backbone for the experiments of our semi-dynamic framework.

Extensive empirical evaluations using the Qwen3 family (0.6B and 4B)~\citep{yang2025qwen3technicalreport} confirm that our density-aware framework consistently outperforms static, fixed-ratio baselines. Notably, our analysis reveals a direct positive correlation between the variance of the dynamically selected ratios and the magnitude of performance improvement over static baselines, definitively proving that our framework's superiority stems directly from its adaptive utilization of text diversity rather than extraneous training artifacts.

In summary, our main contributions are threefold:
\begin{itemize}[noitemsep, topsep=0pt, parsep=0pt, partopsep=0pt]
\item \textbf{Identifying the Continuous Hyperparameter Pitfall:} We expose the structural limitations of fully dynamic compression ratio methods, providing evidence as to why LLMs fail when tasked with infinite variations of input-dependent structural hyperparameters.
\item \textbf{Semi-Dynamic Compression:} We propose a novel compression framework that naturally adapts to text information density via discrete ratio auto-selection, advancing the Pareto frontier of current context compression methods with minimal additional overhead.
\item \textbf{Streamlined Training Pipeline:} We introduce a single-stage, pure-SFT training methodology driven by high-quality, open-sourced synthetic data, making the training of soft context compression models more efficient and reproducible.
\end{itemize}

\section{Related Work}

\paragraph{Hard Prompt Compression}
Hard compression methods, such as LLMLingua~\citep{jiang_llmlingua_2023, pan_llmlingua-2_2024}, operate directly within the discrete text space to prune redundant tokens. While these approaches avoid extensive model training, they are inherently bounded by the discrete nature of the vocabulary, struggling to achieve extreme compression ratios without severe information loss.

\paragraph{Soft Context Compression}
Soft compression maps discrete token sequences into shorter, continuous latent representations. Early explorations like xRAG~\citep{cheng_xrag_2024} and 500xCompressor~\citep{li_500xcompressor_2024} aggressively compressed entire documents into a single token embedding, which inevitably caused massive information loss for lengthy documents. Intermediate methods like ICAE~\citep{ge_-context_2024} and PCC~\citep{dai_pretraining_2025} popularized the ``compression tokens'' paradigm. However, these frameworks typically require massive text-reconstruction pre-training and often freeze the decoder, resulting in semantic misalignment. Methods such as Mean-pooling Context Compression~\citep{feldman_simple_2025} discard heavy pre-training in favor of knowledge distillation. 
Conversely, Cascade Context Compression~\citep{liu_context_2025} utilizes 1 million pages of diverse OCR data (encompassing both Chinese and English documents) alongside text reconstruction tasks for its pre-training phase. Concurrently, approaches like Arcaligner~\citep{li_arcaligner_2026} introduce specialized decoder modules, while CLaRa~\citep{he_clara_2025} utilizes high-quality synthetic data to jointly train the compressor and generator over fixed-length targets.

\paragraph{Dynamic and Adaptive Compression}
While most soft compression techniques enforce rigid ratios, some recent works explore text-adaptive strategies. Dynamic Large Concept Models~\citep{qu_dynamic_2026} attempt to chunk text into semantic concepts based on adjacent-token similarity, subsequently applying mean-pooling to each individual chunk to extract its features. However, its chunking strategy is somewhat heuristic and it lacks a mechanism for user-controlled global compression scaling. Similarly, REFRAG~\citep{lin_refrag_2025} employs a reinforcement learning-trained selector for a binary routing decision (compress entirely or leave uncompressed) for each document block. In contrast, our work introduces a semi-dynamic, continuous-to-discrete selection mechanism that seamlessly adapts to varying densities while providing explicit, continuous control over the global compression scale.

\begin{figure*}[t]
    \centering
    \includegraphics[width=0.6\linewidth]{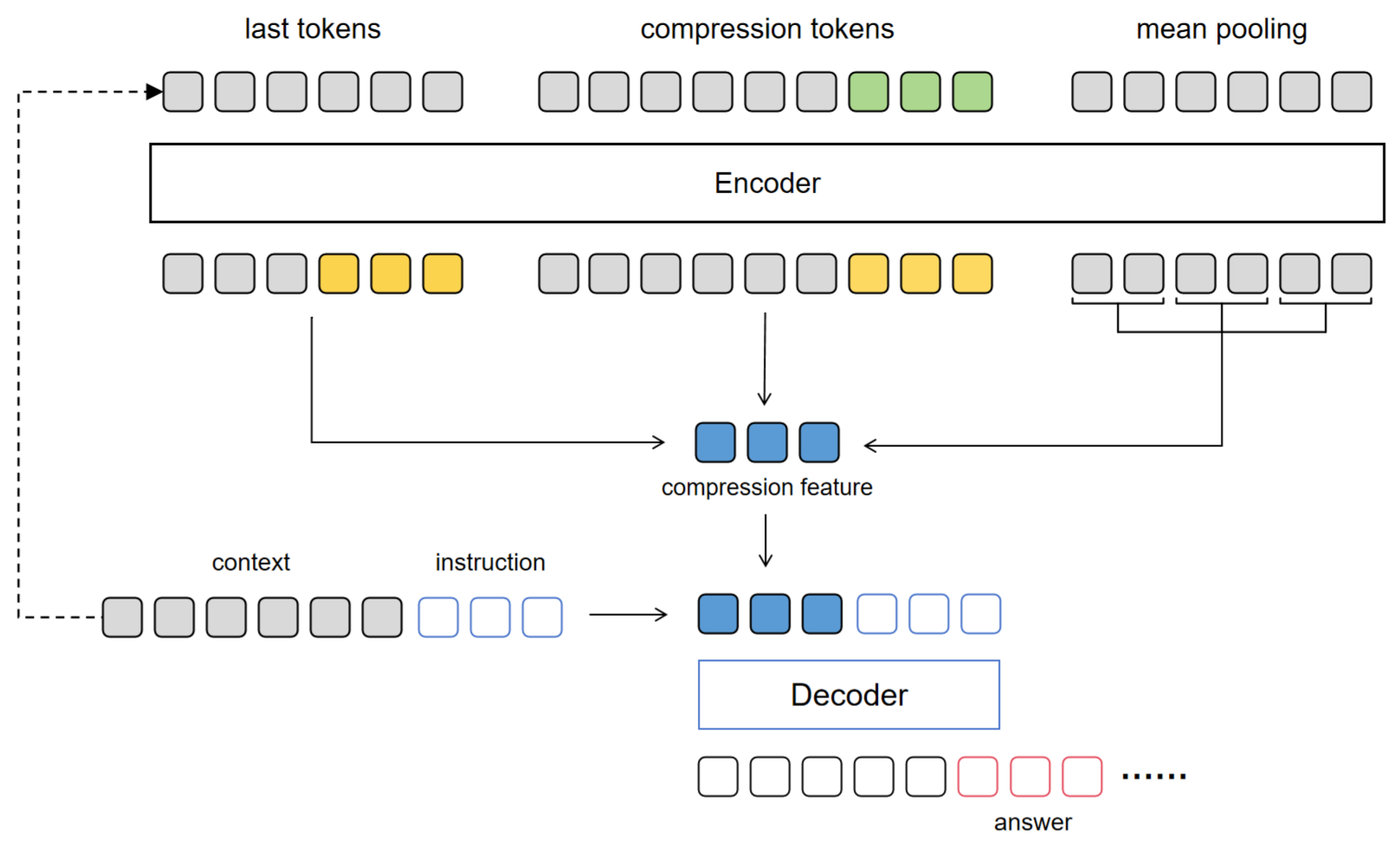}
    \caption{Three typical feature extraction mechanisms for soft context compression.}
    \label{fig:3methods}
\end{figure*}

\begin{figure*}[t]
    \centering
    \includegraphics[width=0.7\linewidth]{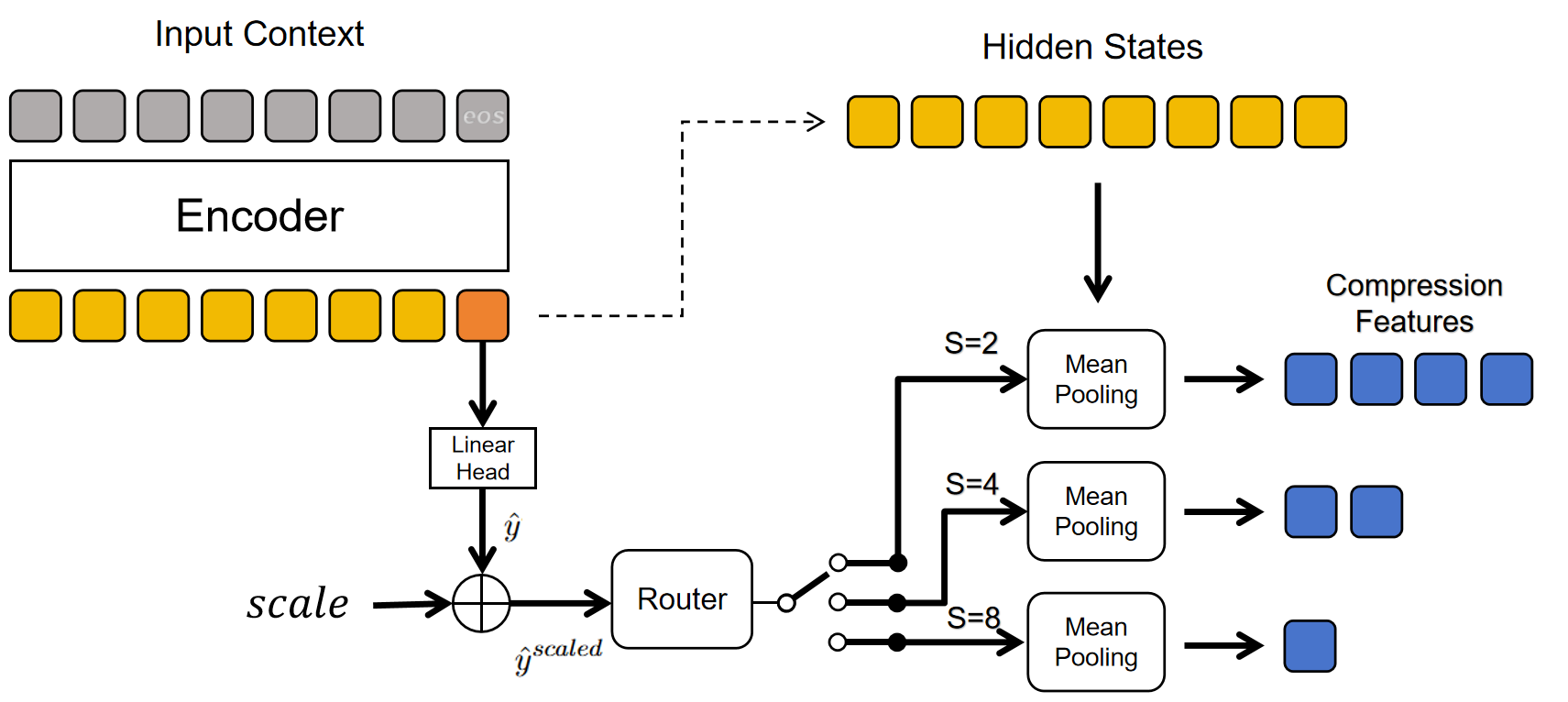}
    \caption{Our semi-dynamic context compression method, utilizing mean-pooling as the optimal structural backbone.}
    \label{fig:semi}
\end{figure*}

\section{Methodology}

To systematically address the context compression bottleneck, we first review existing feature extraction paradigms to explain why a fully dynamic compression ratio would lead to infinite structural hyperparameters. Building upon this, we detail the core failure of fully dynamic compression, which motivates our Semi-Dynamic framework.

\subsection{Re-evaluating Feature Extraction Methods}

Soft context compression relies on a feature extraction mechanism to derive a compressed latent representation from an encoder's hidden states. Given an input context of length $L_{ctx}$, the goal is to extract a representation of reduced length $M$, i.e. $M$ latent tokens. As shown in Figure \ref{fig:3methods}, we categorize existing extraction operations into 3 primary paradigms:

\begin{itemize}[noitemsep, topsep=0pt, parsep=0pt, partopsep=0pt]
    \item \textbf{Last Tokens:} A naive approach that directly extracts the hidden states of the final $M$ tokens of the original sequence. The structural hyperparameter is the target token count $M$.
    \item \textbf{Compression Tokens:} The widely adopted paradigm that appends $M$ learnable tokens to the end of the context for information gathering. After encoding, the hidden states corresponding to these tokens are extracted. The structural hyperparameter is also $M$.
    \item \textbf{Mean-Pooling:} A chunking-free approach that partitions the encoded sequence into non-overlapping windows. By applying mean-pooling over the hidden states within each window, it produces the compressed vectors. The structural hyperparameter here is the pool size $S$.
\end{itemize}

A fundamental tension arises when controlling the compression behavior of these methods. For token-based methods (last tokens and compression tokens), maintaining a specific \textit{compression ratio} $r$ requires the hyperparameter $M$ to be strictly dependent on the input context length (i.e., $M \approx r \cdot L_{ctx}$), or manually split the context into chunks of fixed length. Conversely, for mean-pooling, outputting a fixed \textit{compressed length} $M$ requires the pool size $S$ to be dynamically dependent on $L_{ctx}$ (i.e., $S \approx L_{ctx} / M$). Consequently, without dynamic hyperparameters, token-based methods must be inherently fixed-length, while mean-pooling is inherently fixed-ratio.

\subsection{The Pitfall of Continuous Structural Hyperparameters}
The dependency issues outlined above naturally motivate \textit{fully dynamic} compression, where structural hyperparameters ($M$ or $S$) are dynamically computed to adapt to varying information densities. However, theoretically, LLMs map inputs to fixed computational sub-graphs. When a hyperparameter dictates the structure of the graph, such as dynamically determining the exact number of tokens $M$ to append or the exact stride $S$ of a pooling window as a continuous function of $L_{ctx}$, it creates an infinite spectrum of computational variations, making optimization highly unstable. 

Our empirical investigations confirm this: when models are forced into fully continuous dynamic setups (the "continuous" here is not its mathematical meaning, but refers to a too vast variety of variations, thus can be considered relatively "continuous" in integer space), they suffer severe accuracy degradation. Conversely, training a model simultaneously on a small, discrete set of fixed operations (e.g., $S \in \{2, 4, 8, 16\}$) maintains near-optimal accuracy. This definitive contrast highlights that models can robustly learn a finite set of distinct structural operations, but fail against the infinite variations of a continuous dynamic parameter.

\subsection{The Semi-Dynamic Compression Framework}
Guided by the necessity for finite structural operations, we propose the \textbf{Semi-Dynamic Context Compression} framework (Figure \ref{fig:semi}). It retains the flexibility of density-aware compression while actively avoiding the continuous hyperparameter pitfall. 

\paragraph{Discrete Ratio Selector (DRS)}
To bridge the gap between continuous density prediction and discrete structural execution, we propose \textbf{Discrete Ratio Selector (DRS)}, a rule-based module between the encoder and decoder. At its core, the DRS functions mathematically as a scalar quantizer: it maps a continuous predicted signal into a predefined, finite set of discrete states. 

Initially, the encoder's regression head outputs a continuous value $\hat{y}$, representing the predicted compression ratio in logarithmic space ($\log_2$). To enable zero-shot controllable inference, we introduce a user-defined hyperparameter, $scale$, which acts as an additive bias to the head's prediction:
\begin{equation}
 \hat{y}^{scaled} = \hat{y} + scale
\end{equation}
By adjusting $scale$ at inference, users can smoothly shift the overall distribution toward better fidelity (negative scale) or better efficiency (positive scale). The continuous predicted compression ratio is then recovered via exponentiation:
\begin{equation}
 \hat{r} = 2^{\hat{y}^{scaled}}
\end{equation}

The subsequent quantization process branches based on the chosen structural backbone:

\textbf{Case 1: Ratio-Based Quantization (e.g., Mean-Pooling).}
We define a predefined candidate set of discrete ratios $\mathcal{R} = \{r_1, r_2, \dots, r_k\}$ (e.g., $\{0.125, 0.25, 0.5\}$). The continuous ratio $\hat{r}$ is quantized to the nearest discrete candidate $r_{target}$:
\begin{equation}
 r_{target} = \mathop{\arg\min}_{r \in \mathcal{R}} |\hat{r} - r|
\end{equation}
The discrete pooling window size $S$ is then deterministically computed as $S = \text{int}(1/r_{target})$, ensuring a valid, finite structural operation.

\textbf{Case 2: Length-Based Quantization (e.g., Compression Tokens).}
We define a candidate set of discrete token counts $\mathcal{M} = \{m_1, m_2, \dots, m_k\}$ (e.g., $\{16, 32, 64, 128\}$). For a given context of length $L_{ctx}$, we calculate the continuous target token count $\hat{m}$ and quantize it to the nearest available discrete count $M_{target}$:
\begin{equation}
 \hat{m} = \hat{r} \cdot L_{ctx}
\end{equation}
\begin{equation}
 M_{target} = \mathop{\arg\min}_{m \in \mathcal{M}} |\hat{m} - m|
\end{equation}

By decoupling the continuous density prediction from the discrete structural execution through this DRS quantization, the model operates exclusively within the finite set of structural parameters it can reliably learn.

\paragraph{Single-Stage Architecture and Dynamic Expansion}
To ensure computational efficiency, we designed a single-stage architecture that completes density prediction and compression in a single encoding pass. Given the context, the encoder first produces hidden states $H \in \mathbb{R}^{L_{ctx} \times d}$ ($d$ is hidden size). We extract the hidden state of the final token, $h_{last}$, passing it through a linear regression head to predict the continuous compression target $\hat{y}$. Next, $\hat{y}$ is routed through the Discrete Ratio Selector (DRS) to determine the exact discrete parameter ($r_{target}$ or $M_{target}$). Only after this parameter is selected does the model execute the structural compression over $H$ to extract the condensed representations. Finally, these representations are mapped into the decoder's input embeddings via an MLP projector.

To simplify the user prompt, we introduce \textit{dynamic single-placeholder expansion}. The user inserts only a single placeholder token to replace the original context. When preparing the input for the decoder, this token is dynamically expanded to the required length dictated by $r_{target}$ (or $M_{target}$), and its input embeddings are replaced by the projected compression features.

\subsection{Density-Aware Data Synthesis and Label Generation}

Unlike previous density-aware methods~\citep{lin_refrag_2025} relying on complex Reinforcement Learning (RL) pipelines (like PPO), we propose a pure Supervised Fine-Tuning (SFT) approach driven by synthetic data. This avoids the optimization instabilities inherently associated with RL.

\paragraph{Motivation for the Density Proxy}
Our approach relies on the intuition that \textbf{the length of a highly condensed summary reflects the original text's information density}. While an imprecise heuristic, the discretized nature of our framework means the continuous proxy label does not need flawless precision; it only needs to provide a rough indicator to steer the prediction into the correct discrete bucket.

\paragraph{Dual-Phase Data Synthesis}
We perform synthetic data generation in two phases using a teacher LLM (e.g., Qwen3-30B-A3B-Instruct) on seed contexts from the UltraFineWeb~\citep{wang2025ultrafineweb} dataset. This dataset comprises a robust mixture of bilingual pre-training data, where the English subset is rigorously filtered from Fineweb-v1.4~\citep{penedo2024the} and the Chinese subset is filtered from Chinese-Fineweb-V2~\citep{yu2025opencsgchinesecorpusseries}.

\begin{itemize}[noitemsep, topsep=0pt, parsep=0pt, partopsep=0pt]
\item \textbf{Phase 1: Task Synthesis for Generative Loss.} We generate standard QA pairs and summaries to compute the causal language modeling loss ($\mathcal{L}_{LM}$), jointly optimizing the encoder, projector, and decoder.
\item \textbf{Phase 2: Ultra-Concise Synthesis for Density Labels.} We prompt the teacher LLM to generate extremely concise summaries omitting all redundant words, whose lengths ($L_{sum}$) are used as the intrinsic density proxy for training label creation.
\end{itemize}

\paragraph{Label Formulation and Joint Optimization}
For a context of length $L_{ctx}$ and ultra-concise summary length $L_{sum}$, the target density label in logarithmic space is defined as:
\begin{equation}
 y = \log_2\left(\frac{L_{ctx}}{L_{sum}}\right)
\end{equation}
The logarithmic transformation is critical for optimization stability. Taking the base-2 logarithm ensures that the label distribution remains roughly uniform across a linear space. Without it, as the summary length $L_{sum}$ linearly decreases for highly compressible texts, the raw ratio $\frac{L_{ctx}}{L_{sum}}$ would rapidly expand following an inverse proportional curve. This would result in an heavily skewed target distribution dominated by excessively large label values, leading to inherently biased model predictions. Finally, the joint model is optimized using the LM loss and Mean Squared Error (MSE) for the prediction head:
\begin{equation}
 \mathcal{L}_{total} = \mathcal{L}_{LM} + \lambda \cdot \text{MSE}(\hat{y}, y)
\end{equation}


\section{Experiments}

\subsection{Experimental Setup}

\paragraph{Training Data}
We construct a synthetic dataset of 10 million samples, whose seed contexts are sampling from UltraFineWeb~\citep{wang2025ultrafineweb} with context lengths between 128 and 1,300 tokens. Using Qwen3-30B-A3B-Instruct, we generate context-based NLP tasks encompassing summarization, single/multi-document QA, and multi-hop reasoning in English and Chinese. All of our training experiments are based on this synthetic dataset.

\paragraph{Evaluation Benchmarks}
We construct a mixed dataset for evaluation, from four standard reading comprehension benchmarks (filtered under 2,048 tokens), uniformly sampling 1,000 instances from: HotpotQA~\citep{yang2018hotpotqa}, SQuAD~\citep{rajpurkar-etal-2016-squad}, Natural Questions (NQ)~\citep{47761}, and AdversarialQA~\citep{bartolo2020beat}.

\paragraph{Evaluation Metrics}
We evaluate mainly using 2 metrics: answer accuracy and average compression ratio. For accuracy, we use \textit{substring accuracy}: a score of 1 is awarded if the exact reference answer appears anywhere within the output, which is more intuitive than F1 and more aligns with human assessment than exact-match. For average compression ratio, it is calculated as the sum of the original context lengths of all the samples which are corrected answered divided by the sum of their compressed lengths. Noteworthy, here we apply a strict \textit{validity filter}: count \textbf{only} for instances answered correctly. This prevents the samples that are aggressively compressed by the model but fail to generate correct answers from artificially inflating the average compression ratio.

\subsection{Implementation Details}
We employ the Qwen3 family, initializing the encoder from \texttt{Qwen3-0.6B} and the decoder also from \texttt{Qwen3-0.6B}. For SFT, we apply LoRA ($r=16$, alpha 128 for encoder, 64 for decoder) on all the linear modules with a global batch size of 80. For the discretized mechanism, ratio-based candidate sets are $\mathcal{R} = \{2\times, 4\times, 8\times, 16\times, 32\times\}$. We append an $<eos>$ token to the context to let the encoder know the last token's hidden state is specifically used for ratio prediction. The converter is a 2-layer MLP with the intermediate size of 4,096. For mean-pooling compression, the encoder's attention is turned to bidirectional. 

\subsection{Main Results}

\begin{figure*}[ht]
 \centering
  \includegraphics[width=0.7\linewidth]{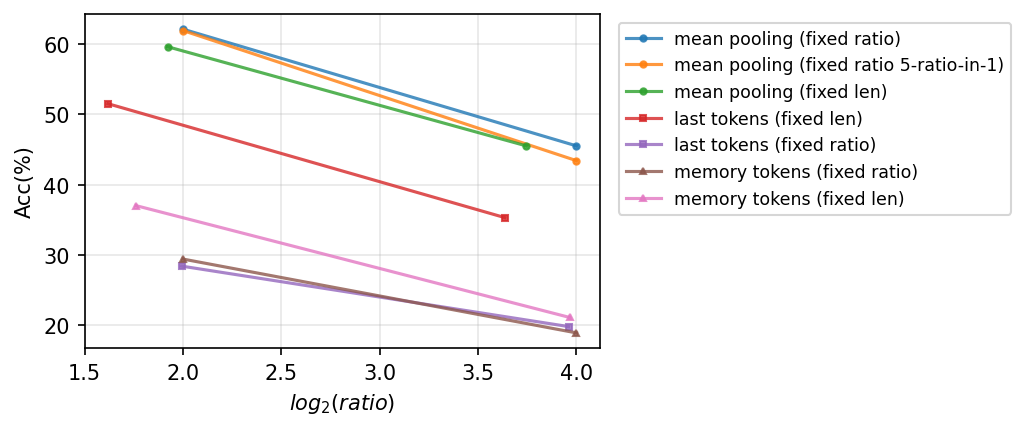} 
 \caption{Accuracy vs. Average Compression Ratio across three feature extraction methods (mean-pooling, last tokens, compression tokens), evaluated under fixed-ratio and fixed-length settings.}
 \label{fig: diff methods}
\end{figure*}

\begin{figure*}[ht]
 \centering
  \includegraphics[width=0.6\linewidth]{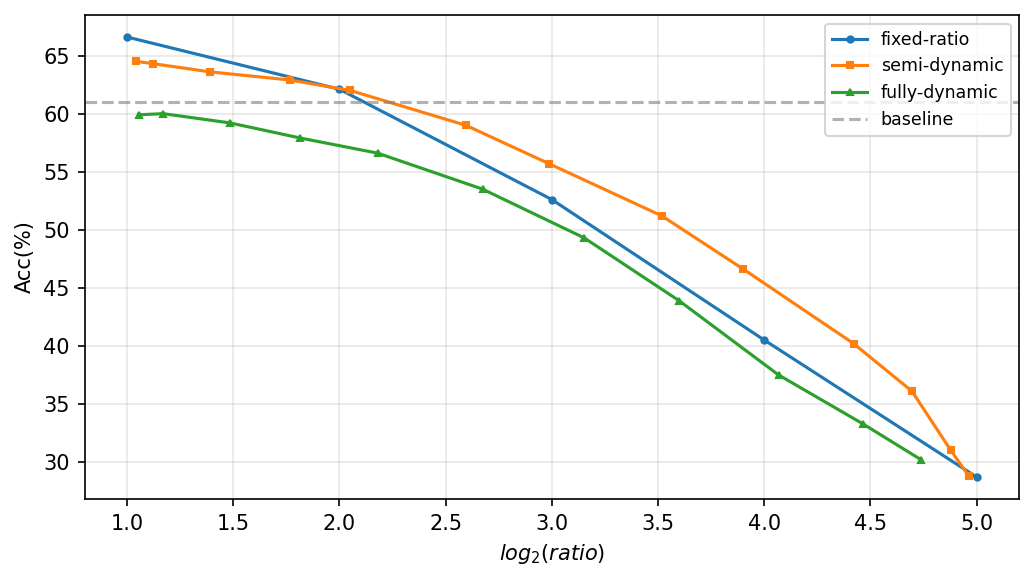} 
 \caption{Accuracy vs. Average Compression Ratio for fixed-ratio vs. semi-dynamic mean-pooling. For the semi-dynamic, the $scale$ parameter is varied across $\{-2, -1.5, -1, -0.5, 0, 0.5, 1, 1.5, 2, 2.5, 3, 3.5, 4\}$ to achieve gradually growing compression ratio. The baseline dashed line represents not using compression.}
 \label{fig: dynamic}
\end{figure*}





\begin{figure}[ht]
    \centering
    \begin{minipage}{0.45\textwidth}
            \centering
        \includegraphics[width=\linewidth]{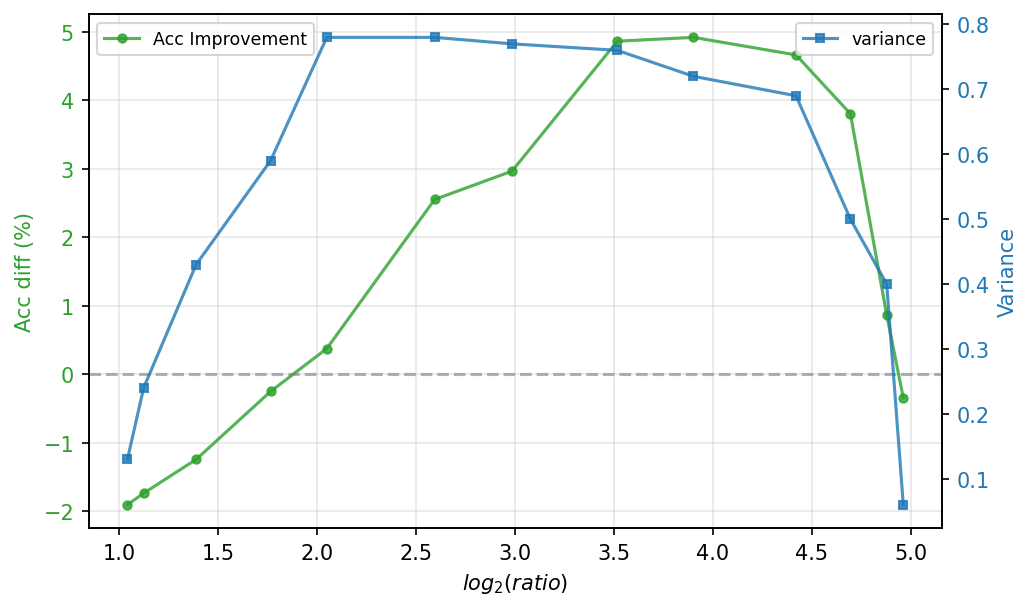}
        \caption{Variance of selected compression ratios ($\log_2$) and absolute accuracy improvement of semi-dynamic method over the fixed-ratio baseline.}
        \label{fig: variance_dynamism}

    \end{minipage}\hfill
    \begin{minipage}{0.45\textwidth}
        \centering
        \includegraphics[width=\linewidth]{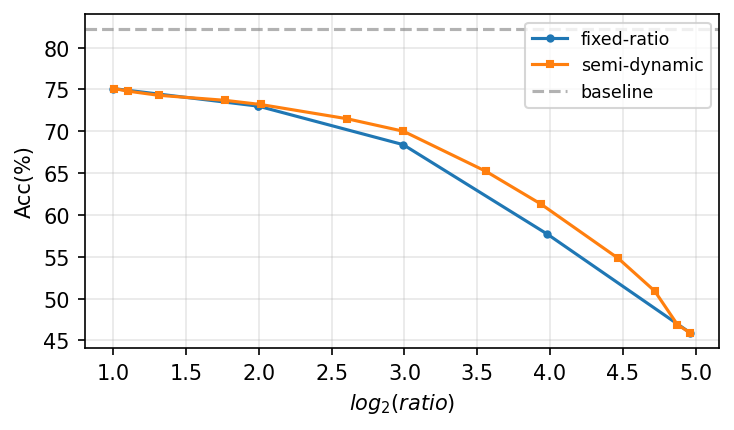}
        \caption{Accuracy vs. Average Compression Ratio for Qwen3-4B-Instruct with fixed-ratio and semi-dynamic mean-pooling compression.}
        \label{fig:4b}
    \end{minipage}
    
    \vspace{1.5em}
    
    \begin{minipage}{0.45\textwidth}
        \centering
        \includegraphics[width=\linewidth]{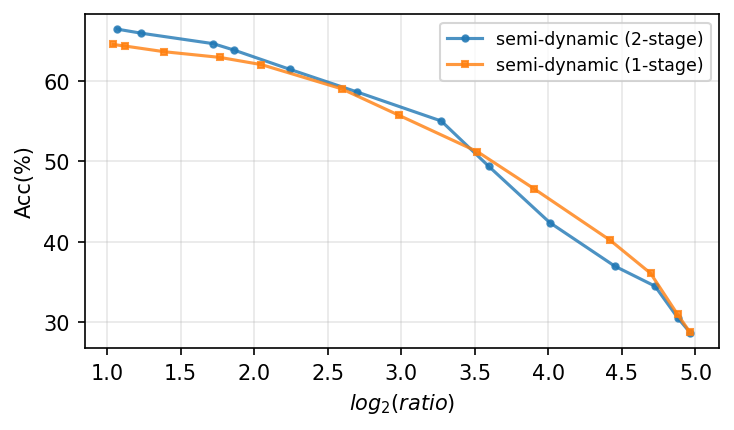}
        \caption{Accuracy vs. Average Compression Ratio for 2-stage and 1-stage semi-dynamic mean-pooling compression.}
        \label{fig:2-stage}
        
    \end{minipage}\hfill
    \begin{minipage}{0.45\textwidth}
        \centering
        \includegraphics[width=\linewidth]{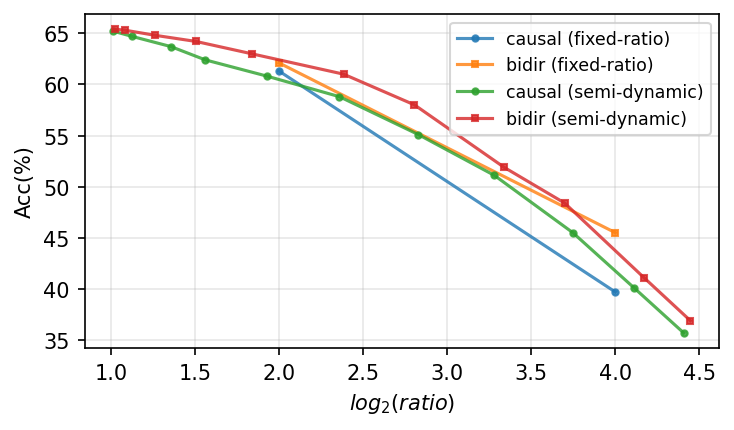}
        \caption{Accuracy vs. Average Compression Ratio for mean-pooling with causal or bidirectional attention.}
        \label{fig: causal}
    \end{minipage}
\end{figure}

\subsubsection{Backbone Comparisons and the Hyperparameter Pitfall}

Our first experiments isolate the architectural mechanisms by training and evaluating with the 3 feature extraction, on 4 different settings: fixing the compression ratio at 4 and 16 respectively, and fixing the compression length to 32 and 128 respectively. The results are shown in
Figure \ref{fig: diff methods}, where each method corresponds to 2 average compression ratios. 

\paragraph{Dominance of Mean-Pooling}
When comparing feature extraction methods, at equivalent average compression rates, we find mean-pooling consistently outperforms both token-based methods. Surprisingly, the widely adopted compression tokens paradigm is significantly outperformed even by naive last tokens extraction. This may be due to the fact that the additional trainable parameters required for compression tokens do not provide additional useful information and are only for the formal rationality of the structure, struggling to converge optimally alongside the rest of the model, unless adopting a heavy pretraining phase.

\paragraph{The Hyperparameter Space Pitfall}
When evaluating fixed-ratio versus fixed-length settings, the empirical data highlights a strong inverse correlation between the size of the structural hyperparameter space and downstream accuracy. Given our training context lengths (128–1,024):
\begin{enumerate}
    \item A \textbf{Fixed Ratio} regime for token-based methods forces the target token count $M$ to fluctuate widely (e.g., between 32 and 256, based on the context length of the training sample), exposing the model to over 200 distinct hyperparameters of operations and causing massive quality drops compared to the fixed-length regime where the hyperparameter is fixed.
    \item A \textbf{Fixed Length} regime for mean-pooling forces the stride $S$ to fluctuate (e.g., 4 to 32), a hyperparameter set of 28 distinct operations, resulting in a slight but noticeable accuracy drop compared to the fixed-ratio regime where the hyperparameter is fixed.
    \item The \textbf{5-Ratio-in-1 Mean-Pooling} setup explicitly constrains the hyperparameter set to just five discrete values ($\mathcal{R} = \{2, 4, 8, 16, 32\}$) (for each training batch, we randomly fix one of these 5 ratios; in evaluation we fix the ratio to 4 and 16 respectively). Because the model multiplexes between only 5 fixed hyperparameters of operations, the accuracy drop is minimal, which is consistent with the findings of Mean-pooling Context Compression~\citep{feldman_simple_2025}.
\end{enumerate}
This confirms that while LLMs can effectively multiplex a small set of discrete structural operations (semi-dynamic), navigating the vast hyperparameter space (fully-dynamic) causes severe optimization failures.

\subsubsection{Semi-Dynamic vs. Fixed-Ratio Compression}

We evaluate our Semi-Dynamic framework against standard fixed-ratio baselines (training an individual model for each ratio) using the mean-pooling backbone (since it is just proved optimal). As the results shown in Figure \ref{fig: dynamic}, we can find that:

\begin{enumerate}
    \item \textbf{Semi-Dynamic Outperforms Static:} At identical average compression ratios, the density-aware semi-dynamic method maintains higher accuracy than static ratio method across the entire evaluated spectrum, except for the lowest ratio. While the fully dynamic method performs even worse than the static ratio method.
    \item \textbf{Maximum Gain at Moderate Scales:} The performance gap is most pronounced at moderate $scale$ biases. As illustrated in Figure \ref{fig: variance_dynamism}, the variance of the model's selected compression ratio peak at moderate average compression ratios (typically ranging from 4 to 16), showing basically the same trend as the accuracy improvements over the fixed-ratio baseline. While at extreme high/low scales, almost all samples are forced into the maximum/minimum discrete bucket, stifling the advantage. This further confirms that our framework's superiority indeed stems directly from its \textbf{dynamic adaptability}, but not more training data or training tricks.

\end{enumerate}

\subsection{Supplementary Experiments}

\paragraph{Scaling to Larger Base Models}
As shown in Figure \ref{fig:4b}, replicating the experiments with \texttt{Qwen3-4B-Instruct} (for initializing both the encoder and decoder) confirms that the 4B models exhibit a significantly higher overall accuracy at any setting. Yet the relative performance gap between static and semi-dynamic methods persists, proving our semi-dynamic framework scales effectively with model capacity.

\paragraph{1-Stage vs. 2-Stage} As shown in Figure \ref{fig:2-stage}, we compare a 2-stage pipeline (using a standalone, isolated regression model for ratio prediction) against our single-stage joint model. The single-stage model's performance is very close to the 2-stage pipeline, indicating that jointly training 2 functions (compression and ratio prediction) into one encoder poses no harm, and achieves higher efficiency.

\paragraph{Bidirectional vs. Causal Attention}
As shown in Figure \ref{fig: causal}, for fixed-ratio mean-pooling compression, comparing a standard causal encoder against a bidirectional encoder (causal mask disabled) reveals negligible differences at low compression ratios ($2\times, 4\times$). While at higher compression ratios ($\ge 16\times$), the bidirectional encoder's global visibility provides a distinct advantage in determining salient features during aggregation. For the semi-dynamic setting, bidirectional encoder is always slightly better than causal encoder. This confirms the necessity of bidirectional attention.

\section{Conclusion}

In this work, we introduce the Semi-Dynamic Context Compression framework to address the core inefficiencies of rigid, fixed-ratio soft compression. By identifying the intrinsic inability of LLMs to optimize over continuous structural hyperparameters, we propose the Discrete Ratio Selector (DRS). This novel quantization mechanism successfully bridges continuous density prediction with discrete, learnable structural execution. Implemented within a highly efficient single-stage architecture, our approach dynamically adapts to the intrinsic information density of varying texts while offering users smooth control over global compression aggressiveness via a simple scaling parameter. Furthermore, we established a streamlined, pure-SFT training pipeline that utilizes summary length as a highly effective density proxy, eliminating the need for complex reinforcement learning or expensive text-reconstruction pre-training. Extensive empirical evaluations confirm that our density-aware framework outperforms static baselines, successfully leveraging text diversity to establish a robust new Pareto frontier for context compression techniques.

\bibliographystyle{colm2026_conference}
\bibliography{colm2026_conference}

@inproceedings{cheng_xrag_2024,
    author = {Xin Cheng and
Xun Wang and
Xingxing Zhang and
Tao Ge and
Si{-}Qing Chen and
Furu Wei and
Huishuai Zhang and
Dongyan Zhao},
    bibsource = {dblp computer science bibliography, https://dblp.org},
    booktitle = {Advances in Neural Information Processing Systems 38: Annual Conference
on Neural Information Processing Systems 2024, NeurIPS 2024, Vancouver,
BC, Canada, December 10 - 15, 2024},
    editor = {Amir Globersons and
Lester Mackey and
Danielle Belgrave and
Angela Fan and
Ulrich Paquet and
Jakub M. Tomczak and
Cheng Zhang},
    title = {xRAG: Extreme Context Compression for Retrieval-augmented Generation
with One Token},
    url = {http://papers.nips.cc/paper\_files/paper/2024/hash/c5cf13bfd3762821ef7607e63ee90075-Abstract-Conference.html},
    year = {2024}
}

@inproceedings{dai_pretraining_2025,
    address = {Vienna, Austria},
    author = {Dai, Yuhong and Lian, Jianxun and Huang, Yitian and Zhang, Wei and Zhou, Mingyang and Wu, Mingqi and Xie, Xing and Liao, Hao},
    booktitle = {Proceedings of the 63rd {Annual} {Meeting} of the {Association} for {Computational} {Linguistics} ({Volume} 1: {Long} {Papers})},
    doi = {10.18653/v1/2025.acl-long.1394},
    editor = {Che, Wanxiang and Nabende, Joyce and Shutova, Ekaterina and Pilehvar, Mohammad Taher},
    isbn = {979-8-89176-251-0},
    language = {en-US},
    note = {TLDR: This paper proposes a decoupled compressor-LLM framework, pre-trained on text reconstruction and completion tasks, designed to effectively preserve essential contextual information within condensed embedding representations to reduce inference costs while preserving the down-stream LLM configurations.},
    pages = {28715--28732},
    publisher = {Association for Computational Linguistics},
    title = {Pretraining {Context} {Compressor} for {Large} {Language} {Models} with {Embedding}-{Based} {Memory}},
    url = {https://aclanthology.org/2025.acl-long.1394/},
    urldate = {2025-12-31},
    year = {2025}
}

@misc{feldman_simple_2025,
    author = {Feldman, Yair and Artzi, Yoav},
    journal = {ArXiv preprint},
    title = {Simple {Context} {Compression}: {Mean}-{Pooling} and {Multi}-{Ratio} {Training}},
    url = {https://arxiv.org/abs/2510.20797},
    volume = {abs/2510.20797},
    year = {2025}
}

@inproceedings{ge_-context_2024,
    author = {Tao Ge and
Jing Hu and
Lei Wang and
Xun Wang and
Si{-}Qing Chen and
Furu Wei},
    bibsource = {dblp computer science bibliography, https://dblp.org},
    booktitle = {Proc. of ICLR},
    publisher = {OpenReview.net},
    title = {In-context Autoencoder for Context Compression in a Large Language
Model},
    url = {https://openreview.net/forum?id=uREj4ZuGJE},
    year = {2024}
}

@misc{he_clara_2025,
    author = {He, Jie and Bai, Richard He and Williamson, Sinead and Pan, Jeff Z. and Jaitly, Navdeep and Zhang, Yizhe},
    journal = {ArXiv preprint},
    title = {{CLaRa}: {Bridging} {Retrieval} and {Generation} with {Continuous} {Latent} {Reasoning}},
    url = {https://arxiv.org/abs/2511.18659},
    volume = {abs/2511.18659},
    year = {2025}
}

@inproceedings{jiang_llmlingua_2023,
    address = {Singapore},
    author = {Jiang, Huiqiang  and
Wu, Qianhui  and
Lin, Chin-Yew  and
Yang, Yuqing  and
Qiu, Lili},
    booktitle = {Proc. of EMNLP},
    doi = {10.18653/v1/2023.emnlp-main.825},
    editor = {Bouamor, Houda  and
Pino, Juan  and
Bali, Kalika},
    pages = {13358--13376},
    publisher = {Association for Computational Linguistics},
    title = {{LLML}ingua: Compressing Prompts for Accelerated Inference of Large Language Models},
    url = {https://aclanthology.org/2023.emnlp-main.825},
    year = {2023}
}

@misc{li_500xcompressor_2024,
    author = {Li, Zongqian and Su, Yixuan and Collier, Nigel},
    journal = {ArXiv preprint},
    title = {{500xCompressor}: {Generalized} {Prompt} {Compression} for {Large} {Language} {Models}},
    url = {https://arxiv.org/abs/2408.03094},
    volume = {abs/2408.03094},
    year = {2024}
}

@misc{li_arcaligner_2026,
    author = {Li, Jianbo and Jiang, Yi and Zhao, Sendong and Hu, Bairui and Wang, Haochun and Qin, Bing},
    journal = {ArXiv preprint},
    title = {{ArcAligner}: {Adaptive} {Recursive} {Aligner} for {Compressed} {Context} {Embeddings} in {RAG}},
    url = {https://arxiv.org/abs/2601.05038},
    volume = {abs/2601.05038},
    year = {2026}
}

@misc{lin_refrag_2025,
    author = {Lin, Xiaoqiang and Ghosh, Aritra and Low, Bryan Kian Hsiang and Shrivastava, Anshumali and Mohan, Vijai},
    journal = {ArXiv preprint},
    title = {{REFRAG}: {Rethinking} {RAG} based {Decoding}},
    url = {https://arxiv.org/abs/2509.01092},
    volume = {abs/2509.01092},
    year = {2025}
}

@misc{liu_context_2025,
    author = {Liu, Fanfan and Qiu, Haibo},
    journal = {ArXiv preprint},
    title = {Context {Cascade} {Compression}: {Exploring} the {Upper} {Limits} of {Text} {Compression}},
    url = {https://arxiv.org/abs/2511.15244},
    volume = {abs/2511.15244},
    year = {2025}
}

@misc{pan_llmlingua-2_2024,
    author = {Pan, Zhuoshi and Wu, Qianhui and Jiang, Huiqiang and Xia, Menglin and Luo, Xufang and Zhang, Jue and Lin, Qingwei and Rühle, Victor and Yang, Yuqing and Lin, Chin-Yew and Zhao, H. Vicky and Qiu, Lili and Zhang, Dongmei},
    journal = {ArXiv preprint},
    title = {{LLMLingua}-2: {Data} {Distillation} for {Efficient} and {Faithful} {Task}-{Agnostic} {Prompt} {Compression}},
    url = {https://arxiv.org/abs/2403.12968},
    volume = {abs/2403.12968},
    year = {2024}
}

@misc{qu_dynamic_2026,
    author = {Qu, Xingwei and Wang, Shaowen and Huang, Zihao and Hua, Kai and Yin, Fan and Zhu, Rui-Jie and Zhou, Jundong and Min, Qiyang and Wang, Zihao and Li, Yizhi and Zhang, Tianyu and Xing, He and Zhang, Zheng and Song, Yuxuan and Zheng, Tianyu and Zeng, Zhiyuan and Lin, Chenghua and Zhang, Ge and Huang, Wenhao},
    journal = {ArXiv preprint},
    title = {Dynamic {Large} {Concept} {Models}: {Latent} {Reasoning} in an {Adaptive} {Semantic} {Space}},
    url = {https://arxiv.org/abs/2512.24617},
    volume = {abs/2512.24617},
    year = {2025}
}

@misc{wang2025ultrafineweb,
    archiveprefix = {arXiv},
    author = {Yudong Wang and Zixuan Fu and Jie Cai and Peijun Tang and Hongya Lyu and Yewei Fang and Zhi Zheng and Jie Zhou and Guoyang Zeng and Chaojun Xiao and Xu Han and Zhiyuan Liu},
    eprint = {2505.05427},
    primaryclass = {cs.CL},
    title = {{Ultra-FineWeb}: Efficient Data Filtering and Verification for High-Quality LLM Training Data},
    year = {2025}
}

@inproceedings{yang2018hotpotqa,
  title={{HotpotQA}: A Dataset for Diverse, Explainable Multi-hop Question Answering},
  author={Yang, Zhilin and Qi, Peng and Zhang, Saizheng and Bengio, Yoshua and Cohen, William W. and Salakhutdinov, Ruslan and Manning, Christopher D.},
  booktitle={Conference on Empirical Methods in Natural Language Processing ({EMNLP})},
  year={2018}
}

@inproceedings{rajpurkar-etal-2016-squad,
    title = "{SQ}u{AD}: 100,000+ Questions for Machine Comprehension of Text",
    author = "Rajpurkar, Pranav  and
      Zhang, Jian  and
      Lopyrev, Konstantin  and
      Liang, Percy",
    editor = "Su, Jian  and
      Duh, Kevin  and
      Carreras, Xavier",
    booktitle = "Proceedings of the 2016 Conference on Empirical Methods in Natural Language Processing",
    month = nov,
    year = "2016",
    address = "Austin, Texas",
    publisher = "Association for Computational Linguistics",
    url = "https://aclanthology.org/D16-1264",
    doi = "10.18653/v1/D16-1264",
    pages = "2383--2392",
    eprint={1606.05250},
    archivePrefix={arXiv},
    primaryClass={cs.CL},
}

@article{47761,
title	= {Natural Questions: a Benchmark for Question Answering Research},
author	= {Tom Kwiatkowski and Jennimaria Palomaki and Olivia Redfield and Michael Collins and Ankur Parikh and Chris Alberti and Danielle Epstein and Illia Polosukhin and Matthew Kelcey and Jacob Devlin and Kenton Lee and Kristina N. Toutanova and Llion Jones and Ming-Wei Chang and Andrew Dai and Jakob Uszkoreit and Quoc Le and Slav Petrov},
year	= {2019},
journal	= {Transactions of the Association of Computational Linguistics}
}

@article{bartolo2020beat,
    author = {Bartolo, Max and Roberts, Alastair and Welbl, Johannes and Riedel, Sebastian and Stenetorp, Pontus},
    title = {Beat the AI: Investigating Adversarial Human Annotation for Reading Comprehension},
    journal = {Transactions of the Association for Computational Linguistics},
    volume = {8},
    number = {},
    pages = {662-678},
    year = {2020},
    doi = {10.1162/tacl\_a\_00338},
    URL = { https://doi.org/10.1162/tacl_a_00338 },
    eprint = { https://doi.org/10.1162/tacl_a_00338 }
}

@inproceedings{
  penedo2024the,
  title={The FineWeb Datasets: Decanting the Web for the Finest Text Data at Scale},
  author={Guilherme Penedo and Hynek Kydl{\'\i}{\v{c}}ek and Loubna Ben allal and Anton Lozhkov and Margaret Mitchell and Colin Raffel and Leandro Von Werra and Thomas Wolf},
  booktitle={The Thirty-eight Conference on Neural Information Processing Systems Datasets and Benchmarks Track},
  year={2024},
  url={https://openreview.net/forum?id=n6SCkn2QaG}
}

@misc{yu2025opencsgchinesecorpusseries,
      title={OpenCSG Chinese Corpus: A Series of High-quality Chinese Datasets for LLM Training}, 
      author={Yijiong Yu and Ziyun Dai and Zekun Wang and Wei Wang and Ran Chen and Ji Pei},
      year={2025},
      eprint={2501.08197},
      archivePrefix={arXiv},
      primaryClass={cs.CL},
      url={https://arxiv.org/abs/2501.08197}, 
}

@misc{yang2025qwen3technicalreport,
    author = {Qwen3 Team},
    journal = {ArXiv preprint},
    title = {Qwen3 Technical Report},
    url = {https://arxiv.org/abs/2505.09388},
    volume = {abs/2505.09388},
    year = {2025}
}

\end{document}